\documentclass{article}
\usepackage{ijcai22}

\usepackage{times}
\usepackage{soul}
\usepackage{url}
\usepackage[hidelinks]{hyperref}
\usepackage[utf8]{inputenc}
\usepackage[small]{caption}
\usepackage{graphicx}
\usepackage{pgfplots}
\pgfplotsset{compat=1.17}
\usepackage{amsmath}
\usepackage{amsthm}
\usepackage{booktabs}
\usepackage{algorithm}
\usepackage{algorithmic}
\usepackage{siunitx}
\usepackage{placeins}

\DeclareMathOperator{\NSD}{NSD}
\DeclareMathOperator{\NMI}{NMI}
\DeclareMathOperator{\NPMI}{NPMI}
\DeclareMathOperator{\OD}{OD}

\title{Assessing Demographic Bias Transfer from Dataset to Model: A Case Study in Facial Expression Recognition}

\author{
Iris Dominguez-Catena\footnote{Contact Author}\and
Daniel Paternain\and
Mikel Galar\\
\affiliations
Institute of Smart Cities (ISC), Department of Statistics, Computer Science and Mathematics\\
Public University of Navarre (UPNA)\\
Arrosadia Campus, 31006, Pamplona, Spain \\
\emails
\{iris.dominguez, daniel.paternain, mikel.galar\}@unavarra.es
}

\begin{document}

\maketitle

\begin{abstract}
The increasing amount of applications of Artificial Intelligence (AI) has led researchers to study the social impact of these technologies and evaluate their fairness. Unfortunately, current fairness metrics are hard to apply in multi-class multi-demographic classification problems, such as Facial Expression Recognition (FER). We propose a new set of metrics to approach these problems. Of the three metrics proposed, two focus on the representational and stereotypical bias of the dataset, and the third one on the residual bias of the trained model. These metrics combined can potentially be used to study and compare diverse bias mitigation methods. We demonstrate the usefulness of the metrics by applying them to a FER problem based on the popular Affectnet dataset. Like many other datasets for FER, Affectnet is a large Internet-sourced dataset with 291,651 labeled images. Obtaining images from the Internet raises some concerns over the fairness of any system trained on this data and its ability to generalize properly to diverse populations. We first analyze the dataset and some variants, finding substantial racial bias and gender stereotypes. We then extract several subsets with different demographic properties and train a model on each one, observing the amount of residual bias in the different setups. We also provide a second analysis on a different dataset, FER+.
\end{abstract}

\section{Introduction}

When algorithms and automated systems interact with users, they can often cause harm in many unintentional ways. This effect is multiplied when the system is a machine learning system trained to imitate human behavior, which is inherently conditioned by prejudice and cognitive biases. In machine learning, the biases gathered in the training information can leak to the trained models, which are otherwise expected to be fair. When these systems are deployed to the real world they have been shown to exhibit gender, racial and other demographic biases \cite{Keyes2018,Prabhu2020,Buolamwini2018}.

The current state-of-the-art systems employ very large datasets, with the most renowned example being Imagenet, with over 14 million images. Further analysis on this dataset has shown critical biases and problematic data \cite{Prabhu2020,Dulhanty2019}. The amount of models trained on this dataset \cite{Denton2021}, which is already a standard for model pretraining, means that any transference of bias from the dataset to the trained models could impact very large populations in unpredictable ways.

This work explores a new metric-based methodology for the analysis of bias in machine learning problems, focusing on the measure of bias transfer between the dataset and the trained model. Despite the definition of multiple fairness metrics \cite{Pessach2020}, to the best of our knowledge there is no bias metric supporting multi-class classification problems studied for multiple potentially protected groups, capable of isolating dataset bias from model bias. In particular, our methodology is oriented to the bias transfer from the dataset to the model, where most mitigation systems can be implemented (given a fixed dataset with its own inherent bias).

Three metrics are proposed in this work. The first two metrics measure representational bias in the dataset. One of them is dedicated to quantifying the representational imbalance, where some demographic groups are over or under-represented in the dataset. The other one is a novel usage of the Normalized Mutual Information (NMI) and Normalized Pointwise Mutual Information (NPMI) metrics \cite{Bouma2009}. We propose using these metrics for stereotype measurement, a type of bias where demographic group representations differ among classes. These two metrics serve as a baseline for measuring the bias present in the input data. Over that baseline, we can then employ a third proposed metric to measure the bias present in the trained model. This metric measures the variation of recall per class among multiple simultaneous demographic groups, giving a single output value quantifying the amount of bias with respect to the different demographic groups. The three metrics combined enable the study of the bias transfer from the dataset to the trained model.

In particular, we apply these metrics to analyze the bias transfer in a Facial Expression Recognition (FER) problem. The objective of FER systems is to identify the facial expression of people in either video or images, in an attempt to detect the underlying emotion. The nature of the data (human faces) in these problems make them prone to diverse biases and misrepresentations. Human biases in this context have already been studied \cite{Elfenbein2002}.

Although the application of these metrics requires demographic information of the subjects in the dataset, the usage of existing demographic models makes recovering some of this information possible for unlabeled datasets. In the context of FER problems we employ a pretrained model, Fairface \cite{Karkkainen2021} to obtain some demographic descriptors, namely apparent race and gender, from the face images. Although these demographic predictions are only an approximation of the real attributes, in the absence of more accurate data they can be used to perform a general analysis.

This work is aimed at helping reduce the racial, ethnic and gender inequalities that can arise in the development and deployment of AI systems. Specifically, the proposed study of demographic bias transfer can be used to both detect demographic bias in machine learning systems and assess the impact of mitigation methods, guiding the efforts in the implementation of fairer systems in general.

Code to replicate the reported results is available at GitHub\footnote{\url{https://github.com/irisdominguez/Dataset-Bias-Metrics}}.

\section{Related Work}

\subsection{Algorithmic Fairness}\label{ssec:fairness}
The concept of algorithmic fairness is usually built around the absence of bias or harm. In particular, several independent bias taxonomies have been proposed, focused on different definitions and aspects of bias. Taxonomies like that of \cite{Suresh2021} consider up to 7 sources of bias in a machine learning pipeline, split in two subgroups: 
\begin{itemize}
    \item Biases in the data generation, comprising historical bias, representational bias and measurement bias.
    \item Biases in the building and implementation of the system, comprising learning bias, aggregation bias, evaluation bias and deployment bias.
\end{itemize}

This work focuses on the measurement of bias in both subgroups independently, and more specifically, in the source dataset and the trained model. The measurement of bias in the source dataset aggregates historical and representational biases, and is common to any model and deployment that employs the same dataset. The measurement of model bias includes learning and aggregation biases. Jointly measuring both biases will give us an estimation of the amount of bias that leaks from the dataset to the model. This can help guide our efforts to the most critical parts of the system, easing the study of mitigation strategies. The remaining forms of bias are not directly applicable to our analysis.

\subsubsection{Bias metrics}\label{ssec:metrics}
There are multiple fairness and bias metrics defined through the literature \cite{Pessach2020}. They are commonly defined for binary classification problems and two populations, a general population and a protected group, with one or several protected attributes indicating the membership to these populations.

However, in many applications there are multiple demographic groups that could require protection at the same time, and more than two target classes, with none of them being clearly advantageous. To the best of our knowledge, there is no bias metric covering this casuistry that is also able to isolate true model bias from validation dataset imbalance. Therefore, we focus on multi-class systems, studying bias across multiple demographic groups. Our bias definition considers any differential treatment suffered by any of the demographic groups in any of the problem classes. To design our metrics, we take the following as reference.

\textbf{Disparate impact} \cite{Feldman2015} asserts that the proportion of predictions of the positive class is similar across groups: \begin{equation}
    \frac{P(\hat{y}=1|s\neq1)}{P(\hat{y}=1|s=1)}\geq 1-\varepsilon\ ,
\end{equation}
where $s$ is the protected attribute ($1$ for the privileged group and $\neq 1$ otherwise) and $\hat{y}$ the class prediction ($1$ for the positive class and $\neq 1$ otherwise).

Disparate Impact is arguably the most common fairness metric. However, it requires a defined positive outcome and privileged and protected groups, making it unfit for our application.

\textbf{Overall accuracy equality} \cite{Berk2018} requires a similar accuracy across groups: 
\begin{equation}
    \left| P(y=\hat{y}|s=1)- P(y=\hat{y}|s\neq1) \right| \leq \varepsilon\ ,
\end{equation}
where $s$ is the protected attribute ($1$ for the privileged group and $\neq 1$ otherwise), $\hat{y}$ the predicted class and $y$ the real class.

As it focuses on the accuracy, the Overall accuracy equality metric is applicable in both binary and multi-class problems. It also treats all the target classes equally, without requiring one of them to be defined as positive or advantageous. Unfortunately, it still requires defined privileged and protected groups, making it suitable only for the analysis of individual demographic groups.

\textbf{Mutual Information} \cite{Kamishima2012} measures the statistical dependence between two attributes: 
\begin{equation}
    \sum_{\hat{y} \in \hat{Y}} \sum_{s \in {S}} P(\hat{y},s)\log\frac{P(\hat{y},s)}{P(\hat{y})P(s)}\leq\varepsilon\ ,
\end{equation}
where $s$ denotes the protected attribute (from a set $S$) and $\hat{y}$ denotes the prediction (from a set $\hat{Y}$).

Mutual Information is one of the few fairness metrics that can be directly applied to multi-class classification problems, even when considering several potentially protected groups. Despite this, the metric only measures the dependence between the protected attributes and either the predicted or real class. In problems where the validation partition is not balanced (the real class and protected attributes are dependent), the Mutual Information measured over the trained model prediction becomes unable to disentangle that imbalance from the potential model bias.

Nonetheless, for our analysis we propose the employment of two variants of the Mutual Information metric to measure the bias not in the final model, but in the source dataset, where they only consider the real class and the protected attributes. These variants are the Normalized Mutual Information (NMI) and the Normalized Pointwise Mutual Information (NPMI) proposed by \cite{Bouma2009} in the context of collocation extraction. They are both bounded and easier to interpret than the classical Mutual Information, and in the case of the $\NPMI$, it can also detect specific stereotypes on top of the general dataset bias. The mathematical definitions are given in Section~\ref{ssec:dataset_metric_nmi}.

\subsection{Facial Expression Recognition}\label{ssec:fer}
The problem of automatic FER is commonly used as a proxy to the more general emotion recognition. Although many works raise questions about the universality of facial expressions to convey emotions across cultures \cite{Elfenbein2002,Jack2009}, and despite the developments in other emotion measurement modalities \cite{Gonzalez-Sanchez2017}, FER is still one of the most used methods. The applications of these systems are multiple, ranging from robotics \cite{Sonmez2021} to assistive technology \cite{Joseph2021}.

Most works deal with either a continuous emotion codification, such as the Pleasure-Arousal-Dominance model \cite{Mehrabian1996} or a discrete codification, such as the six/seven basic emotions proposed by \cite{Ekman1971}. This work focuses on the second approach, the most used in modern machine learning.

\section{Proposal}

\subsection{Representational Bias Metric}\label{ssec:dataset_metric_nsd}

The most basic level of demographic analysis that can be performed in a dataset is that of representational bias, where there is unequal representation of different demographic groups in the overall dataset. A clear predominance of a demographic group in the dataset can hint at a potential bias in favor of that group, generating a differentiated and potentially harmful behavior of the resulting model. Previous bias metrics (presented in Section~\ref{ssec:metrics}) only measure the bias for the final model, so they cannot detect this specific bias. As a way to measure the representational bias of the dataset, we propose using a metric based on the standard deviation of the normalized demographic distribution. This \textbf{Normalized Standard Deviation} (NSD) adjusts the standard deviation of a normalized vector:
\begin{equation}
    \NSD(x) = \frac{n}{\sqrt{n-1}}\sqrt{\frac{\sum_{i=1}^{n}(x_i - \bar{x})^2}{n}}\ ,
\end{equation}
where $n$ is the number of elements of the vector $x$ and $\overline{x}$ stands for the arithmetic mean.

The $\NSD$ calculated over a normalized demographic distribution is bounded in the interval $[0, 1]$, where 0 is no bias (uniform distribution) and 1 is total bias (the full population belongs to a single group).

\subsection{Stereotypical Bias Metric}\label{ssec:dataset_metric_nmi}

A second level of analysis of the dataset not covered by previous metrics is that of the presence of stereotypes, understood as a variation in the demographic profile of each target class in the dataset. In the context of FER, this can result in a different prior for an emotion label depending on the perceived demographic group of the sample.

For most FER datasets, and in general for most multi-class multi-demographic datasets, the analysis and quantification of stereotypical bias is complex due to the double class and demographic imbalance usually accepted in these datasets. To decouple this secondary bias from the main representational bias, we employ the $\NPMI$ metric proposed by \cite{Bouma2009}.

The \textbf{Normalized Pointwise Mutual Information} ($\NPMI$) measures the statistical dependence between two attributes for a specific pair of values:
\begin{equation}
    \NPMI(s, y) = -\frac{\ln\frac{P(s,y)}{P(s)P(y)}} {\ln P(s,y)}\ ,
\end{equation}
where $s$ denotes the protected attribute and $y$ denotes the class. $\NPMI$ values lay in the range $[-1, 1]$, with $1$ being total correlation (overrepresentation), $0$ being no correlation and $-1$ being inverse correlation (underrepresentation).

For an aggregated value representing the summary of the $\NPMI$ biases, we employ the $\NMI$ metric, also proposed by \cite{Bouma2009}.

The \textbf{Normalized Mutual Information} ($\NMI$) measures the statistical dependence between two attributes:
\begin{equation}
    \NMI(S, Y) = -\frac{ \sum_{y \in Y} \sum_{s \in S} P(s,y) \ln\frac{P(s,y)}{P(s)P(y)}} {\sum_{y \in Y} \sum_{s \in S} P(s,y) \ln{P(s,y)}}\ ,
\end{equation}
where $S$ denotes all the demographic groups studied and $Y$ the set of classes of the problem. The $\NMI$ value lies in the range $[0, 1]$, with $0$ being no bias and 1 being total bias. 

\subsection{Model Bias Metric}\label{ssec:model_metric}

Based on the metrics presented in Section~\ref{ssec:metrics}, we require a new metric to measure the model bias in problems like FER, where we operate over multiple demographic groups and multiple classes, with an unbalanced test dataset. For the calculation of the model bias metric, we expect the model to perform with similar recall for each demographic group and for each target class to be considered fair. The \textbf{Recall} $\text{R}(y,s)$ for a class $y$ and a demographic group $s$ is defined as
\begin{equation}
    \text{R}(y, s) = P(\hat{y}=y|y,s)\ .
\end{equation}

It is important to note that each of the classes in the problem can have different inherent difficulties. Therefore, we first calculate the intraclass disparity for each class by aggregating the recalls of that class for all demographic groups. Later, we aggregate those intraclass disparities into a final dataset metric. The \textbf{Intraclass Disparity} (ID) for each class $y$, aggregated for $S$ demographic groups, is defined as:
\begin{equation}
    \text{ID}(y) = \frac{1}{n-1}\sum_{s \in S} \left(
            {1-\frac{\displaystyle \text{R}(y, s)}{\displaystyle \max_{s'\in S}{\text{R}(y, s')}}}\right)\ ,
\end{equation}
with the convention that $\text{ID}(y) = 0$ if $\displaystyle \max_{s'\in S}{\text{R}(y, s')} = 0$.

This metric uses the maximum recall $\text{R}(y,s)$ of any demographic group ($s$) for the class ($y$) as the baseline to obtain a relative value between 0 (same performance as the maximum recall group) and 1 (recall 0 relative to the maximum recall group) for each demographic group. These values are then aggregated, obtaining the final metric. This measure considers all the groups, maximizing the $\text{ID}$ metric to 1 when all groups except the one with the highest recall have an accuracy of 0 (situation of maximum privilege or bias).

Finally, we can study the class disparities by themselves, or aggregate them with a simple mean over the set of classes $C$, giving us an \textbf{Overall Disparity} $\OD$ for the model, defined as:
\begin{equation}
    \OD = \frac{1}{|C|}\sum_{c\in C}{\text{ID}(c)}\ .
\end{equation}

The $\OD$ value has a lower bound of $0$ (no bias) and an upper bound of $1$ (maximum bias). The same bounds apply to the individual ID scores.

\section{Case Study}

\subsection{Dataset}\label{ssec:dataset}

Affectnet \cite{Mollahosseini2019} is a large FER dataset, composed of 420,299 images of facial expressions. $291,651$ of these images are classified in basic emotions, namely \textit{neutral}, \textit{happy}, \textit{sad}, \textit{surprise}, \textit{fear}, \textit{disgust}, \textit{angry} and \textit{contempt}. The dataset is divided into two partitions, one for training and one for validation, with $287,651$ and $4,000$ images respectively.

The dataset was created from Internet image searches, based on queries related to both the emotional labels considered and gender, age, and ethnicity descriptors. The search query strings were also translated to 6 different languages.

The final images have $425$ by $425$ pixels of resolution.

Appendix \ref{appendix:ferplus} repeats the same analysis for a second dataset, FER+ \cite{Barsoum2016}.

\subsection{Determination of Demographic Labels} \label{ssec:fairface}

Most large FER datasets \cite{Benitez-Quiroz2016,Mollahosseini2019} are gathered from regular Internet image searches, with relatively low data curation. Outside FER problems, many datasets have already undergone heavy bias analysis \cite{Dulhanty2019}, but the lack of meta-information and demographic labels on subjects for in the wild (ITW) FER datasets has hindered the bias analysis in this context. Affectnet is not an exception, and despite the diversity considerations during its generation, demographics labels are not available.

In recent years, the development of new datasets for demographic annotation such as Fairface \cite{Karkkainen2021} has enabled the demographic relabeling of existing datasets. It is important to note that this kind of annotation is highly subjective and imperfect, and any demographic label obtained constitutes only a proxy measure for the real demographic characteristics of the subjects, which is already a subjective and complex concept. For example, the gender classification of Fairface is binary (Male, Female), which already constitutes a bias against non-binary people \cite{Keyes2018}. The race classification only uses the seven most common descriptors (White, Latino/Hispanic, East Asian, Black, Middle Eastern, Indian, and Southeast Asian). Furthermore, both categories are treated as single-label classification. Despite these issues, the Fairface model gives us an approximation of the demographic profile of the dataset, enough for a general bias analysis in the absence of more accurate data.

\subsection{Experimental Setup}\label{ssec:experimental}

We employ a simple VGG11 \cite{Simonyan2015} network with no pretraining as the base test model. This is a classical convolutional architecture often used as a baseline for machine learning applications.

The experiments are developed in PyTorch 1.8.0 and Fastai 2.3.1. The hardware employed is a machine equipped with a GeForce RTX 2080 Ti GPU, 128~GB of RAM, an Intel\textregistered\ Xeon\textregistered\ Silver 4214 CPU, and running CentOS Linux 7.7.

All the models are trained under the same conditions and hyperparameters, namely, a maximum learning rate of $1e^{-2}$ with a 1cycle policy (as described in \cite{Smith2018} and implemented in Fastai) for $100$ iterations. This parameter was decided by using the \textit{lr\_finder} tool in Fastai. The batch size is set to $256$, the maximum allowed by the hardware setup. For each dataset, we train the model $10$ times, and average the results over them.

We have also applied basic data augmentation provided by Fastai through the \textit{aug\_transforms} method, including left-right flipping, warping, rotation, zoom, brightness, and contrast alterations.

\subsection{Experiments}\label{ssec:experiments}

\subsubsection{Dataset bias}\label{ssec:exp_presentation_dataset}

To analyze the dataset, first we process it using the Fairface model to obtain a demographic estimation for each image. We then proceed to calculate the demographic representation profile of the dataset, and compute the $\NSD$ bias metric presented in Section~\ref{ssec:dataset_metric_nsd}. After that, we can also use the $\NPMI$ metric to highlight any stereotypical bias inherent to the distribution of labels for each demographic group, as explained in Section~\ref{ssec:dataset_metric_nmi}.

The demographic information added to the dataset through this relabeling process also enables the creation of derived datasets that can be used to simulate different bias situations. In particular, we generate:

\begin{itemize}
    \item Two \textit{balanced subsets} for the racial and gender demographics. These datasets have the same representation of each demographic group considered for each of the target labels. This balancing removes both representational and stereotypical biases.
    \item Two \textit{artificially biased gender subsets}, that contain examples of only one gender.
\end{itemize}

The two gender biased and the gender balanced subsets are generated with exactly the same number of examples for each class to enable the comparison between the three of them.

\subsubsection{Model bias}

All the trained models are evaluated on the whole Affectnet validation partition, and the $\OD$ metric described in Section~\ref{ssec:model_metric} is employed as a measure of bias.

In addition to the subsets proposed in the previous experiment, a series of \textit{stratified subsets} are generated to evaluate the influence of the dataset size in the residual bias of the model. Each subset will contain a fixed percentage of the original train partition, maintaining the same demographic distribution. Hence, they will preserve the dataset biases of Affectnet.

In summary, we want to study the behavior of the model bias metric when increasing the number of training examples and when training on balanced subsets. If balancing the dataset is an efficient mitigation strategy, we expect a lower metric with little to no impact on general accuracy, for the same training data size.

\section{Results And Discussion}

\subsection{Dataset bias}\label{ssec:res_dataset}
\begin{figure}[htb!]
    \centering
    \input{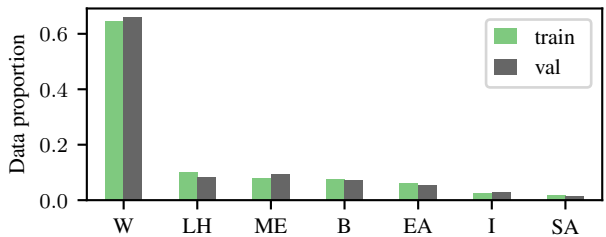}
    \caption{Apparent race distribution of Affectnet (W: White, LH: Latino / Hispanic, ME: Middle Eastern, B: Black, EA: East Asian, I: Indian, SA: Southeast Asian).}\label{figure:source_race_distribution}
\end{figure}

\begin{figure}[htb!]
    \centering
    \input{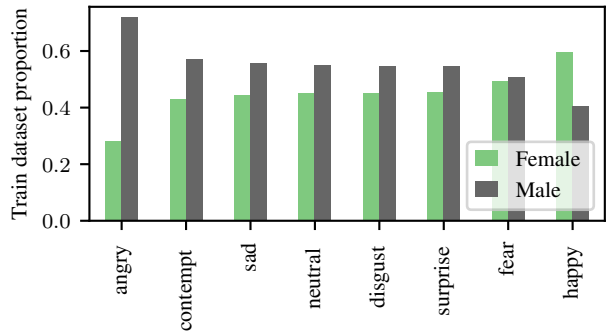}
    \caption{Apparent gender distribution of Affectnet for each label.}\label{figure:source_gender_distribution}
\end{figure}

Figure~\ref{figure:source_race_distribution} shows the apparent race distribution of the dataset in Affectnet. We can observe a heavy imbalance in favor of the white race, which comprises $64.4\%$ of the training data. The apparent gender distribution (not shown) is much more balanced, with $49.7\%$ of the training data classified as male and the rest as female. In the case of gender, imbalances arise in the per-label analysis, summarized in Figure~\ref{figure:source_gender_distribution}, reaching a $72\% - 28\%$ imbalance in the case of the \textit{angry} class.

These intuitive indicators of bias are also reflected in the $\NSD$ and $\NMI$ metrics results presented in Table~\ref{tab:dataset_bias}, where both the original dataset and some variations are analyzed. The representational bias is measured with the $\NSD$ metric and the stereotypical bias with the $\NMI$. Note that the $\NSD$ and $\NMI$ metrics do not share the same scales and are not comparable to each other. The number in parentheses denotes the number of demographic groups considered, 2 for gender and 7 for race. These results show a zero representational and stereotypical bias for the artificially balanced datasets in their respective categories, as expected. Note that the gender biased datasets are excluded from the gender bias calculations, as they only include examples of one gender. For the original dataset, we can observe a relatively low representational bias in the gender category ($\NSD=0.0057$) compared to the race category ($\NSD=0.5902$). On the contrary, the stereotypical bias is higher in the gender category ($\NMI=0.0089$) compared to the race category ($\NMI=0.0021$). This agrees with the imbalance perceived in Figure~\ref{figure:source_gender_distribution}.

\begin{table}[htb!]
\resizebox{\columnwidth}{!}{
\begin{tabular}{@{}llrrrr@{}}
\toprule
              & & \multicolumn{2}{l}{Representational} & \multicolumn{2}{l}{Stereotypical} \\
              & & \multicolumn{2}{l}{bias (NSD)} & \multicolumn{2}{l}{bias (NMI)} \\
\cmidrule{3-6}
      Dataset & &                    Race (7) & Gender (2) &                 Race (7) & Gender (2) \\
\midrule
     Original &  &                    $0.5902$ &   $0.0057$ &                 $0.0021$ &   $0.0089$ \\ \cmidrule{1-6}
     Balanced &     Race &                    \pmb{$0.0000$} &   $0.0159$ &                 \pmb{$0.0000$} &   $0.0091$ \\
              &   Gender &                    $0.5929$ &   \pmb{$0.0000$} &                 $0.0017$ &   \pmb{$0.0000$} \\ \cmidrule{1-6}
Gender biased &        M &                    $0.5702$ &      $-$ &                 $0.0020$ &   $-$ \\
              &        F &                    $0.6129$ &      $-$ &                 $0.0020$ &   $-$ \\
\bottomrule
\end{tabular}}
\caption{Representational ($\NSD$) and stereotypical ($\NMI$) bias metrics for the original dataset and the considered subsets.}\label{tab:dataset_bias}
\end{table}

\begin{figure}[htb!]
    \graphicspath{{images/affectnet/}}
    \input{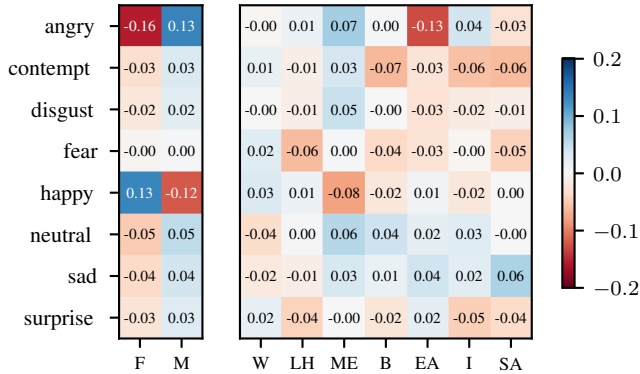}
    \caption{$\NPMI$ analysis of Affectnet. (F: Female, M: Male, W: White, LH: Latino / Hispanic, ME: Middle Eastern, B: Black, EA: East Asian, I: Indian, SA: Southeast Asian)}\label{figure:source_npmi}
\end{figure}

The stereotypical bias detected through the $\NMI$ can be further analyzed with $\NPMI$. Figure~\ref{figure:source_npmi} shows the $\NPMI$ matrices for both the apparent gender and race analysis. In the race category, the most prominent value indicates an underrepresentation ($-0.13$) of the East Asian group in the \textit{angry} class. In the gender category two stereotypical biases stand out: for the \textit{angry} class, an underrepresentation of people recognized as female (and a corresponding overrepresentation of people recognized as male), and the opposite for the \textit{happy} class, with an overrepresentation of people recognized as female and underrepresentation of people recognized as male. The results for the gender category are consistent with the \textit{angry-men-happy-women} social bias already known in the literature \cite{Atkinson2005}.

\subsection{Model bias}\label{ssec:res_model}
\begin{figure}[htb!]
    \centering
    \input{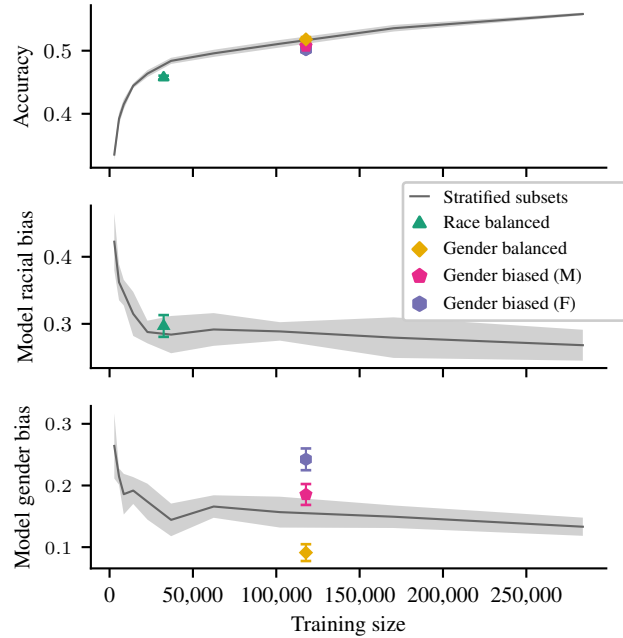}
    \caption{Accuracy and apparent race and gender biases for models.}\label{figure:model_racialgender}
\end{figure}

The results regarding the trained model bias metric $\OD$ and accuracy are shown in Table~\ref{tab:model_bias}, and graphically in Figure~\ref{figure:model_racialgender}.

Regarding racial bias, across the stratified subsets we can observe consistently high model bias values, decreasing clearly from the highest bias in the smallest dataset ($0.422\pm0.043$) to the lowest in the largest dataset ($0.268\pm0.023$). The race balanced dataset, when compared to an imbalanced one of similar size (Stratified / $0.13$), does not seem to improve the bias metric ($0.297\pm0.016$ vs. $0.284\pm0.017$), while decreasing accuracy ($45.7\pm0.3$ vs. $48.4\pm0.5$). In this case, increasing the dataset size (Stratified / $0.22$ and higher) improves the accuracy without impacting the racial bias metric (stable around $0.29$ and decreasing for the larger sizes). Note that the accuracy is calculated over the whole test partition of Affectnet, which is not balanced in terms of labels and the demographic groups studied.

In the gender category, we observe comparatively lower bias values overall, from $0.264\pm0.053$ to $0.133\pm0.015$, coherent with the lower gender bias of the original dataset ($\NSD = 0.0057$, $\NMI = 0.0089$). Balancing the gender does not significantly improve the accuracy results for a similar sized dataset (Stratified / $0.36$, accuracy $51.8\pm0.4$ vs. $51.1\pm0.6$), but in this case the gender bias value improves significantly ($0.091\pm0.014$, lower than all other models). Additionally, although the artificially gender biased datasets have a similar accuracy to the balanced one ($50.7\pm0.5$ and $50.2\pm0.4$ vs. $51.8\pm0.4$), their bias metric is substantially higher ($0.185\pm0.017$ and $0.242\pm0.018$ vs. $0.091\pm0.014$).

\begin{table}[htb!]
\resizebox{\columnwidth}{!}{
\begin{tabular}{@{}llrrll@{}}
\toprule
\multicolumn{4}{l}{}  & \multicolumn{2}{l}{Bias} \\
\cmidrule{5-6}
       \multicolumn{2}{@{}l}{Train data} &   Size &     Accuracy &         Race OD &       Gender OD \\
\midrule
     Original &        1\% &   2,839 & $33.5\pm0.9$ & $0.422\pm0.043$ & $0.264\pm0.053$ \\
              &        2\% &   5,678 & $39.2\pm0.9$ & $0.362\pm0.027$ & $0.214\pm0.013$ \\
              &        3\% &   8,517 & $41.5\pm0.8$ & $0.347\pm0.019$ & $0.186\pm0.033$ \\
              &        5\% &  14,195 & $44.4\pm0.4$ & $0.315\pm0.033$ & $0.192\pm0.022$ \\
              &        8\% &  22,712 & $46.4\pm0.5$ & $0.288\pm0.017$ & $0.174\pm0.029$ \\
              &       13\% &  36,907 & $48.4\pm0.5$ & $0.284\pm0.028$ & $0.144\pm0.026$ \\
              &       22\% &  62,458 & $49.6\pm0.6$ & $0.292\pm0.024$ & $0.166\pm0.018$ \\
              &       36\% & 102,204 & $51.1\pm0.6$ & $0.289\pm0.014$ & $0.157\pm0.025$ \\
              &       60\% & 170,340 & $53.6\pm0.5$ & $0.279\pm0.030$ & $0.149\pm0.018$ \\
              &      100\% & 283,901 & \pmb{$55.8\pm0.2$} & \pmb{$0.268\pm0.023$} & $0.133\pm0.015$ \\ 
\cmidrule{1-6}
     Balanced &       Race &  32,452 & $45.7\pm0.3$ & $0.297\pm0.016$ & $0.177\pm0.017$ \\
              &     Gender & 117,790 & $51.8\pm0.4$ & $0.273\pm0.026$ & \pmb{$0.091\pm0.014$} \\
\cmidrule{1-6}
Gender biased &          M & 117,790 & $50.7\pm0.5$ & $0.277\pm0.015$ & $0.185\pm0.017$ \\
              &          F & 117,790 & $50.2\pm0.4$ & $0.315\pm0.022$ & $0.242\pm0.018$ \\
\bottomrule
\end{tabular}}
\caption{Bias metric summary for the model when trained on dataset variations.}\label{tab:model_bias}
\end{table}

\subsection{Bias transfer}\label{ssec:res_transfer}

Regarding the apparent race analysis of the system, the proposed metrics reveal an important representational imbalance of the dataset coupled with some stereotypical bias. Despite this, when comparing a stratified subset to a dataset of the same size but balanced by race, we observe no improvement in the trained model bias. This suggests that either the source of the model bias is not the dataset bias (inherent differences between racial expressions, for example), or that directly balancing the dataset is not a successful mitigation approach in this context (because of quality differences between the images associated with each race, for example).

Regarding the gender bias, the metrics reveal a comparatively lower representational bias, but a much more pronounced stereotypical bias. In this case, balancing the dataset seems to substantially improve the trained model bias, mitigating the bias transfer. This suggests that the stereotypical bias detected in the dataset has a large impact in the trained model, but the balancing of the dataset corrects it properly.

Additionally, we observe a strong tendency to reduce the bias scores as the size of the training dataset increases, even when the datasets have the same representational and stereotypical biases.

Although the metrics have unveiled both gender and racial bias in the source dataset, these bias transference results suggest that dataset gender bias has a greater impact in the final model. Thus, dataset gender bias seems more susceptible to mitigation measures in the early stages of the AI life-cycle, whereas racial bias may require different mitigation measures in later stages. Further studies would be required to evaluate the impact of different mitigation techniques in this case, but are out of the scope of this paper.

\section{Conclusion}
The metrics presented have been shown to be useful, reflecting some of the biases present in both real and manipulated datasets through easily interpretable values. The analysis of these metrics allows the study of the bias transfer from dataset to trained model, which can be useful for understanding the bias in different stages of a machine learning pipeline, and consequently in the study of mitigation strategies.

In our case study, we have revealed the heavy racial representational bias of a popular FER dataset, Affectnet, and the presence of stereotypical gender biases. The experiments also show how the resulting model seems almost invariant to the removal of the racial bias, while being severely impacted by any gender bias, either induced or corrected, in the source dataset. In the Appendix to this document the results for a second dataset, FER+, are provided, showing similar tendencies to the ones found for Affectnet. These results, while specific to this model and training setup, expose the complexity of the bias analysis and its impact in real world problems.

As future work lines, the same analysis could be performed for more datasets, models, and training setups. Further work in the development of new demographic datasets and models could also improve the accuracy and detail of this bias analysis, and extend it to other problems. Different mitigation techniques, specially in the dataset preprocessing stage, could differ wildly in their impact. Finally, the metrics still require further analysis on their properties and potential application to other contexts. For example, although our proposed $\OD$ reflects both representational and stereotypical biases, having metrics capable of decoupling them could enable a more in-depth bias analysis.

Furthermore, new application areas could be researched in other multi-class multi-demographic AI systems, such as age, gender and race recognition, AI-based medical diagnosis and sign language gesture recognition, to name a few.

\section*{Acknowledgments}
This work was funded by a predoctoral fellowship of the Research Service of Universidad Publica de Navarra, the Spanish MICIN (PID2019-108392GB-I00 and PID2020-118014RB-I00 / AEI / 10.13039/501100011033), and the Government of Navarre (0011-1411-2020-000079 - Emotional Films).

\FloatBarrier
\bibliographystyle{named}
\bibliography{main}

\appendix

\section{FER+}\label{appendix:ferplus}

For completeness, we repeat the analysis on a second dataset, FER+ \cite{Barsoum2016}. FER+ is based on FER2013 \cite{Goodfellow2013}, a large ITW dataset composed of 32,298 color images. FER+ keeps the original images from FER2013, removing only those not containing human faces or of uncertain expressions. The rest of the images are relabeled to reduce the noisiness in the original dataset. This relabeling process was based on 10 independent crowdsourced annotations for each image. The final FER+ dataset consists of 29,397 images.

\subsection{Methodology}

We employ the same experimental setup previously described for Affectnet in Section~\ref{ssec:experimental}.

\subsection{Dataset bias results}

\begin{figure}[htb!]
    \centering
    \input{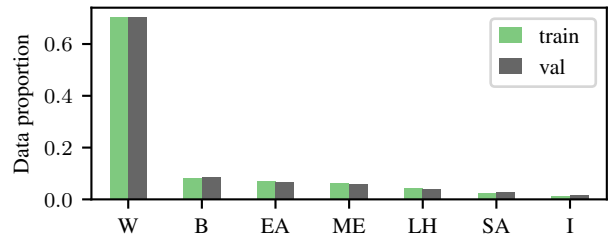}
    \caption{Apparent race distribution of FER+ (W: White, LH: Latino / Hispanic, ME: Middle Eastern, B: Black, EA: East Asian, I: Indian, SA: Southeast Asian).}\label{figure:fp_source_race_distribution}
\end{figure}

\begin{figure}[htb!]
    \centering
    \input{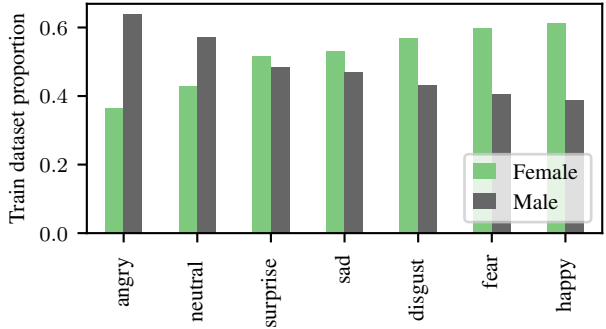}
    \caption{Apparent gender distribution of FER+ for each label.}\label{figure:fp_source_gender_distribution}
\end{figure}

Figure~\ref{figure:fp_source_race_distribution} shows the apparent race distribution of the FER+ dataset. We can observe a heavy imbalance in favor of the white race, which comprises $70.5\%$ of the training data. The apparent gender distribution (not shown) is much more balanced, with $49.9\%$ of the training data classified as male and the rest as female. In the case of gender, imbalances arise in the per-label analysis, summarized in Figure~\ref{figure:fp_source_gender_distribution}, reaching a $63\% - 37\%$ imbalance in the case of the \textit{angry} class. 

The $\NSD$ and $\NMI$ metrics results are presented in Table~\ref{tab:fp_dataset_bias}. Recall that the representational bias is measured here with the $\NSD$ metric and the stereotypical bias with the $\NMI$, and that the two metrics do not share the same scales. The artificially balanced datasets show zero representational and stereotypical bias in their respective categories. The gender biased datasets are excluded from the gender bias calculations as they only include examples of one gender. For the original dataset, we can observe a relatively low representational bias in the gender category ($\NSD=0.0021$) compared to the race category ($\NSD=0.6591$). On the contrary, the stereotypical bias is higher in the gender category ($\NMI=0.0067$) compared to the race category ($\NMI=0.0056$). This agrees with the imbalance perceived in Figure~\ref{figure:fp_source_gender_distribution}.

An important detail of these results is that directly balancing the race in this dataset has an extreme impact both on the representational gender bias ($\NSD=0.0447$ vs. $\NSD=0.0021$ in the original dataset) and the stereotypical gender bias ($\NMI=0.0131$ vs. $\NMI=0.0067$ in the original dataset).

\begin{table}[htb!]
\resizebox{\columnwidth}{!}{
\begin{tabular}{@{}llrrrr@{}}
\toprule
              & & \multicolumn{2}{l}{Representational} & \multicolumn{2}{l}{Stereotypical} \\
              & & \multicolumn{2}{l}{bias (NSD)} & \multicolumn{2}{l}{bias (NMI)} \\
\cmidrule{3-6}
      Dataset & &                    Race (7) & Gender (2) &                 Race (7) & Gender (2) \\
\midrule
     Original &        &                    $0.6591$ &   $0.0021$ &                 $0.0056$ &   $0.0067$ \\ \cmidrule{1-6}
     Balanced &   Race &              \pmb{$0.0000$} &   $0.0447$ &           \pmb{$0.0000$} &   $0.0131$ \\
              & Gender &                $0.6603$ & \pmb{$0.0000$} &             $0.0060$ & \pmb{$0.0000$} \\ \cmidrule{1-6}
Gender biased &      M &                    $0.6413$ &        $-$ &                 $0.0069$ &   $-$ \\
              &      F &                    $0.6868$ &        $-$ &                 $0.0057$ &   $-$ \\
\bottomrule
\end{tabular}}
\caption{Representational ($\NSD$) and stereotypical ($\NMI$) bias metrics for the original dataset and the considered subsets.}\label{tab:fp_dataset_bias}
\end{table}

\begin{figure}[htb!]
    \graphicspath{{images/ferplus/}}
    \input{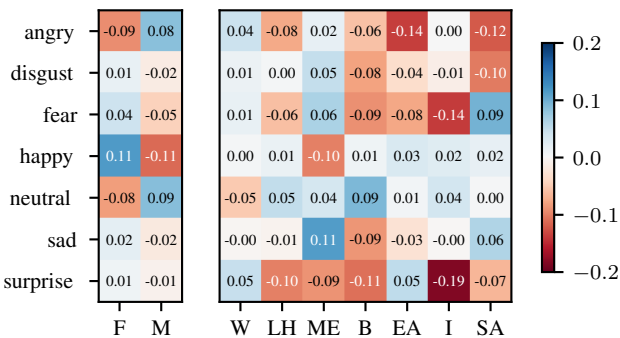}
    \caption{$\NPMI$ analysis of FER+. (F: Female, M: Male, W: White, LH: Latino / Hispanic, ME: Middle Eastern, B: Black, EA: East Asian, I: Indian, SA: Southeast Asian)}\label{figure:fp_source_npmi}
\end{figure}

Figure~\ref{figure:fp_source_npmi} shows the $\NPMI$ matrices for both the apparent gender and race analysis, enabling a more detailed analysis of the $\NMI$ results. In the race category, we can observe more varied and prominent values than what was found in the analysis of Affectnet. In particular, the highest absolute $\NPMI$ value indicates an underrepresentation ($-0.19$) of the Indian group in the \textit{surprise} class. The East Asian group in the \textit{angry} class, the most impacted group for Affectnet, is also heavily underrepresented in FER+, with a $\NPMI$ value of $-0.14$ (the same result was $-0.13$ for Affectnet).

In the gender category two stereotypical biases stand out: for the \textit{angry} class, an underrepresentation of people recognized as female (and a corresponding overrepresentation of people recognized as male), and the opposite for the \textit{happy} class, with an overrepresentation of people recognized as female and underrepresentation of people recognized as male. These two biases were also found in Affectnet. Additionally, we observe a slight overrepresentation of people recognized as female in the rest of the categories, except for the \textit{neutral} class. These results for the gender category are also consistent with the \textit{angry-men-happy-women} social bias already known in the literature \cite{Atkinson2005}.

\subsection{Model bias results}
\begin{figure}[htb!]
    \centering
    \input{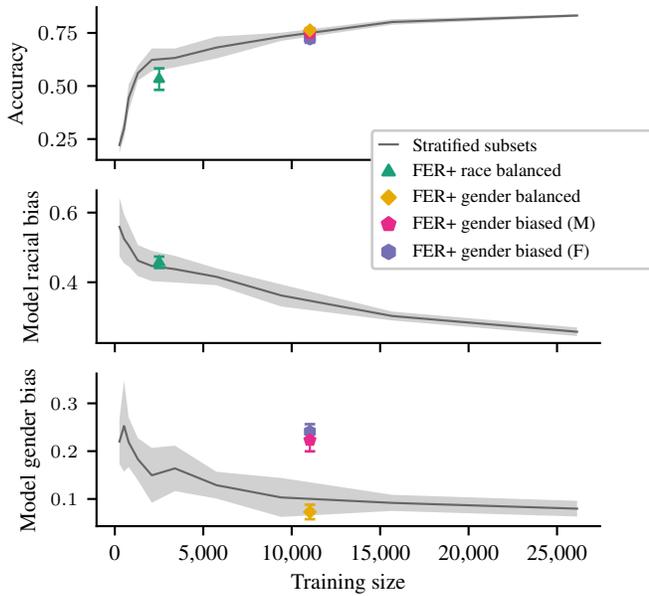}
    \caption{Accuracy and apparent race and gender biases for models.}\label{figure:fp_model_racialgender}
\end{figure}

The results for the trained model bias metric $\OD$ and accuracy are shown in Table~\ref{tab:fp_model_bias}, and graphically in Figure~\ref{figure:fp_model_racialgender}.

The results show a similar behavior as the one previously shown for Affectnet. The main difference is a higher general accuracy (top accuracy $83.2\%$), consistent with the results of \cite{Barsoum2016} for this dataset.

Regarding racial bias, across the stratified subsets we can observe consistently high model bias values, decreasing clearly from the highest bias in the smallest dataset ($0.559\pm0.085$) to the lowest in the largest dataset ($0.258\pm0.012$). The race balanced dataset, when compared to an imbalanced one of similar size (Stratified / $8\%$), does not seem to improve the bias metric ($0.457\pm0.016$ vs. $0.447\pm0.017$), while decreasing accuracy ($53.2\pm5.1$ vs. $62.3\pm5.4$). Larger dataset sizes (Stratified / $8\%$ and higher) improve both the accuracy and the racial bias metric. Note that the accuracy is calculated over the whole test partition of FER+, which is not balanced in terms of labels and the demographic groups studied.

In the gender category, we observe comparatively lower bias values overall, from $0.220\pm0.047$ to $0.080\pm0.016$, coherent with the lower gender bias of the original dataset ($\NSD = 0.0021$, $\NMI = 0.0067$). Balancing the dataset in the gender category does not significantly improve the accuracy results for a similar sized dataset (Stratified / $36\%$, accuracy $73.2\pm1.9$ vs. $76.2\pm0.8$), but the gender bias value improves significantly ($0.073\pm0.015$, lower than all other models). Additionally, although the artificially gender biased datasets have a similar accuracy to the balanced one ($74.7\pm2.5$ and $72.3\pm1.9$ vs. $76.2\pm0.8$), their bias metric is substantially higher ($0.223\pm0.023$ and $0.241\pm0.016$ vs. $0.073\pm0.015$).

\begin{table}[htb!]
\resizebox{\columnwidth}{!}{
\begin{tabular}{@{}llrrll@{}}
\toprule
\multicolumn{4}{l}{}  & \multicolumn{2}{l}{Bias} \\
\cmidrule{5-6}
       \multicolumn{2}{@{}l}{Train data} &   Size &     Accuracy &         Race OD &       Gender OD \\
\midrule
     Original &        1\% &    261 & $22.2\pm3.7$ & $0.559\pm0.085$ & $0.220\pm0.047$ \\
              &        2\% &    522 & $29.9\pm4.3$ & $0.525\pm0.070$ & $0.253\pm0.095$ \\
              &        3\% &    784 & $44.4\pm6.3$ & $0.506\pm0.060$ & $0.219\pm0.052$ \\
              &        5\% &  1,306 & $56.1\pm3.5$ & $0.462\pm0.045$ & $0.183\pm0.044$ \\
              &        8\% &  2,090 & $62.3\pm5.4$ & $0.447\pm0.044$ & $0.149\pm0.057$ \\
              &       13\% &  3,397 & $63.2\pm4.5$ & $0.438\pm0.038$ & $0.164\pm0.048$ \\
              &       22\% &  5,749 & $68.1\pm5.1$ & $0.416\pm0.024$ & $0.129\pm0.028$ \\
              &       36\% &  9,408 & $73.2\pm1.9$ & $0.362\pm0.032$ & $0.103\pm0.041$ \\
              &       60\% & 15,681 & $80.2\pm1.3$ & $0.303\pm0.013$ & $0.092\pm0.017$ \\
              &      100\% & 26,135 & \pmb{$83.2\pm0.6$} & \pmb{$0.258\pm0.012$} & $0.080\pm0.016$ \\
\cmidrule{1-6}
     Balanced &       Race &  2,506 & $53.2\pm5.1$ & $0.457\pm0.016$ & $0.191\pm0.041$ \\
              &     Gender & 11,028 & $76.2\pm0.8$ & $0.368\pm0.011$ & \pmb{$0.073\pm0.015$} \\
\cmidrule{1-6}
Gender biased &          M & 11,028 & $74.7\pm2.5$ & $0.396\pm0.028$ & $0.223\pm0.023$ \\
              &          F & 11,028 & $72.3\pm1.9$ & $0.343\pm0.010$ & $0.241\pm0.016$ \\
\bottomrule
\end{tabular}}
\caption{Bias metric summary for the model when trained on dataset variations.}\label{tab:fp_model_bias}
\end{table}

\subsection{Bias transfer results}

As with Affectnet, the metrics reveal both representational and stereotypical biases, both in race and gender. In the race category, trying to artificially balance FER+ does not improve the trained model bias. For gender, balancing the dataset greatly improves the trained model bias.

The fact that the models behave in a biased way for race, despite artificially balancing the dataset, suggest that the origin of the race bias is not directly linked to the demographic composition of the original dataset. It could be linked to the quality of the images for different races, or to inherent differences in the expression of the people from different racial origins.

Regarding gender, we observe how the bias transference is heavily impacted by bias in the training dataset. This highlights the importance of removing gender stereotypes from the training data as a way of bias mitigation.

Additionally, there is a strong tendency to reduce the bias scores as the size of the training dataset increases, even with stratified datasets that have the same representational and stereotypical biases.

\end{document}